\documentclass[lettersize,journal]{IEEEtran}
\usepackage{graphicx}
\usepackage[caption=false,font=footnotesize]{subfig}
\usepackage{cite}
\usepackage{url}
\usepackage{acronym}
\usepackage{balance}
\usepackage{needspace}
\usepackage{xcolor}
\usepackage{tikz}
\usepackage{enumitem}
\definecolor{ndtblue}{HTML}{527CB5}
\definecolor{plangreen}{HTML}{579D53}
\definecolor{cldaorange}{HTML}{D99A29}

\newcommand{\circledstep}[2]{%
    \tikz[baseline=(char.base)]{
        \node[
            circle,
            fill=#1,
            text=white,
            font=\sffamily\bfseries\scriptsize,
            minimum size=1em,
            inner sep=1pt
        ] (char) {#2};
    }%
}

\begin{document}
\acrodef{AI}{Artificial Intelligence}
\acrodef{NDT}{Network Digital Twin}
\acrodefplural{NDT}[NDTs]{Network Digital Twins}
\acrodef{CDT}{Cognitive Digital Twin}
\acrodefplural{CDT}[CDTs]{Cognitive Digital Twins}
\acrodef{CLDA}{Closed-Loop Dependency Assurance}
\acrodef{LDG}{Loop Dependency Graph}
\acrodef{LLM}{Large Language Model}
\acrodefplural{LLM}[LLMs]{Large Language Models}
\acrodef{PN}{Physical Network}

\title{Can Agentic AI Undermine Its Own Assurance Loop? Closed-Loop Dependency Assurance for Cognitive Network Digital Twins}

\author{Ahmed~Hussain, Konstantinos~Kalogiannis, Salahuddin~Salahuddin, Hexu~Li, and Thiemo~Voigt%
\thanks{Ahmed Hussain and Konstantinos Kalogiannis are with the KTH Royal Institute of Technology, Stockholm, Sweden (e-mail: \{ahmhus, konkal\}@kth.se). Salahuddin Salahuddin is with SSH Communications Security, Helsinki, Finland (e-mail: salahuddin.salahuddin@ssh.com). Hexu Li is with the Tech Validation--AI Department, Chief Technology Officer (CTO) Organization, Lenovo, Beijing, China (e-mail: lihx36@lenovo.com). Thiemo Voigt is with Uppsala University and RISE, Sweden (e-mail: thiemo.voigt@angstrom.uu.se).}
}

\maketitle

\begin{abstract}
Agentic \acf{AI} and \acfp{NDT} are increasingly combined to form \acfp{CDT} that translate operational intents into network changes and support closed-loop management. However, a network change may achieve its intended objective while disrupting the monitoring or recovery mechanisms needed for subsequent decisions. For example, a service may be rerouted while its latency-monitoring traffic continues along the original path, producing measurements that no longer reflect the service's performance. We propose \acf{CLDA}, an assurance mechanism that augments agentic \acs{AI}-enabled \acsp{NDT} with dependency-aware validation. In our framework, an agentic \acs{AI} planner generates and refines network plans, while the \acs{NDT} evaluates the projected network outcomes of these plans. \acs{CLDA} extends this analysis using a \acf{LDG} that connects service requirements to monitoring mechanisms, forwarding paths, failure domains, and recovery prerequisites. \acs{CLDA} projects candidate actions onto these dependencies, checks intermediate states of multi-step plans, and supports deterministic make-before-break repairs using temporary monitoring mechanisms. Unlike the \acf{LLM}-based planner, \acs{CLDA} uses explicit, deterministic rules for assurance decisions and repair. We evaluate \acs{CLDA} through controlled reconfigurations, a $240$-trial \acs{LLM}-based planning campaign, and live network execution. A separate synthetic study examines deterministic assurance scalability across network models representing up to $1{,}024$ routers.
\end{abstract}


\acresetall
\section{Introduction}
\label{sec:intro}

\acp{NDT} allow network operators to evaluate proposed changes before applying them to a physical network~\cite{li2025native,huang2025gdt,huang2026ran}. Through \emph{what-if analysis}, an \acs{NDT} predicts how a change may affect connectivity, routing, and network performance. Meanwhile, \acp{LLM} enable agentic \ac{AI} planners to translate operator requests into network configuration plans and refine them using feedback~\cite{mekrache2026intentweave}. Combining these capabilities moves \acsp{NDT} toward \acp{CDT} that support intent-driven planning and closed-loop network management~\cite{zhou2026ndt,wu2026agentndt}. However, a change that achieves its intended objective may also disrupt the monitoring or recovery mechanisms needed to manage the network afterward.

Consider an operator who asks an agentic \acs{AI} planner to reroute a service between two networks while maintaining connectivity. The planner generates the required configuration changes, and the \acs{NDT} evaluates their expected effects. The service is moved to the new route, but the monitoring traffic used to measure its round-trip latency continues along the original route. The monitor still produces measurements, potentially indicating good performance, even though those measurements no longer represent the path taken by the service. Such measurements constitute \emph{monitoring evidence}, information used to determine whether a service meets its requirements. In this case, the evidence remains available but is no longer valid for assessing the rerouted service.

We call this problem \emph{self-decoupling}: a network change achieves its immediate objective but invalidates a dependency needed for subsequent monitoring or recovery. Similar problems can arise during multi-step reconfigurations. A plan may reach an acceptable final configuration while temporarily leaving the service without valid monitoring between actions. Separately, removing an old route may prevent a registered rollback procedure from restoring the previous configuration. Although these problems are not specific to \acs{AI}, they become particularly important when network changes are planned and executed with limited human intervention.

Existing approaches address aspects of reliable network management. Service-assurance graphs support service-health assessment and fault diagnosis~\cite{claise2023sain}. Network verification checks forwarding and policy invariants~\cite{khurshid2013veriflow}, while consistent-update techniques reason about intermediate configurations~\cite{reitblatt2012updates}. Network-twin, intent-assurance, and agent-supervision approaches support planning and oversight~\cite{ma2026icon,gharbaoui2026assurance}. These approaches provide complementary capabilities, but their integration into an agentic \acs{NDT} control loop must also account for how network changes affect subsequent monitoring and recovery.

To address this gap, we propose \emph{\ac{CLDA}}, a deterministic assurance mechanism that augments agentic \acs{AI}-enabled \acsp{NDT}. A network change should not be authorized solely because the \acs{NDT} predicts an acceptable outcome. The monitoring mechanisms used to verify the service must also remain relevant, and the resources required for recovery must remain available. \acs{CLDA} makes these dependencies explicit through a \ac{LDG} and projects candidate actions onto them to assess whether they remain valid throughout multi-step changes. This allows agentic \acs{AI} to propose and refine network plans while authorization relies on independent, deterministic checks.

\begin{figure*}[!t]
    \centering
    \includegraphics[width=\textwidth]{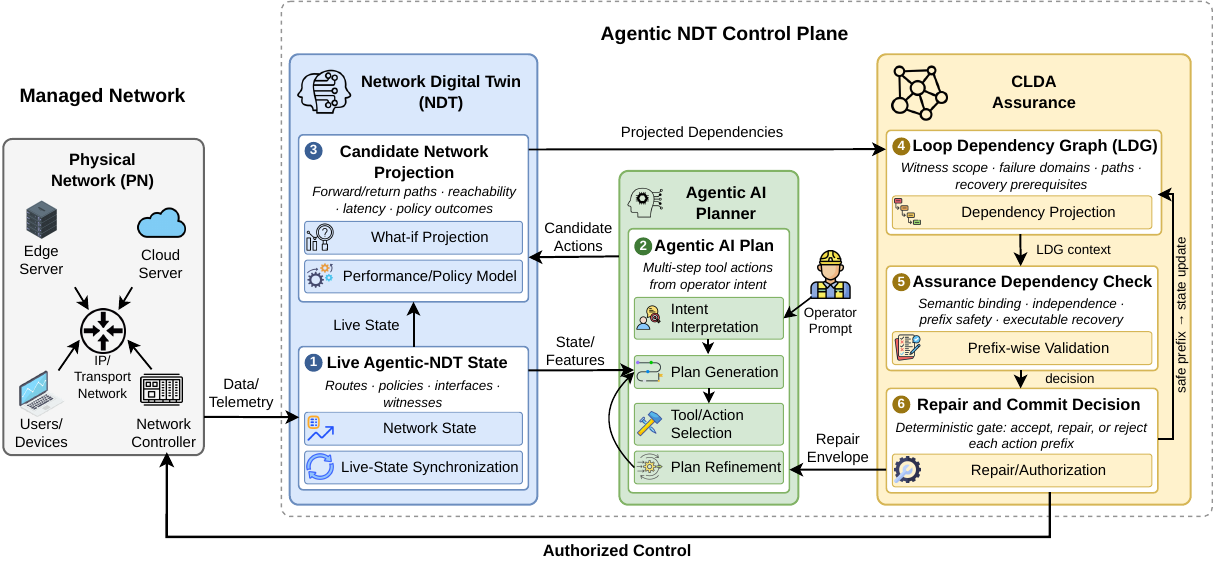}
    \caption{\acs{CLDA}-enhanced agentic \acs{NDT} architecture and assurance workflow. The agentic \acs{AI} planner generates network plans from operator intents, the \acs{NDT} projects their network effects, and \acs{CLDA} evaluates assurance dependencies before accepting, repairing, or rejecting proposed changes.}
    \label{fig:architecture}
\end{figure*}

The main contributions of this work are as follows: (i) We identify self-decoupling as a risk in closed-loop network management and introduce \acs{CLDA}, which represents monitoring and recovery dependencies through an \acs{LDG}. (ii) We integrate dependency checks into pre-commit and execution-time validation, including checks after intermediate actions and deterministic make-before-break repair for supported monitoring gaps. (iii) Our evaluation through controlled network changes, 240 \acs{LLM}-based planning trials, and live execution shows that \acs{CLDA} detects assurance violations and repairs temporary monitoring gaps. A separate synthetic study demonstrates near-linear scaling of deterministic assurance checks up to $1{,}024$ modeled routers.

The remainder of this article is organized as follows. Sec.~\ref{sec:architecture} presents the \acs{CLDA} architecture and \acs{LDG}. Sec.~\ref{sec:assurance} explains dependency checks and repair. Sec.~\ref{sec:case_study} describes the case study and evaluation. Sec.~\ref{sec:open_issues} discusses open research issues, and Sec.~\ref{sec:conclusion} concludes.

\section{\acs{CLDA}-Enhanced Agentic \acs{NDT}}
\label{sec:architecture}

Fig.~\ref{fig:architecture} illustrates the proposed architecture and its assurance workflow. The \ac{PN} provides operational data and telemetry to the \acs{NDT}, which maintains the current network state \circledstep{ndtblue}{1}. Using this information and the operator's intent, the agentic \acs{AI} planner generates and refines candidate network plans \circledstep{plangreen}{2}. The \acs{NDT} performs what-if analysis to project their effects on forwarding paths, reachability, performance, and network policies before execution \circledstep{ndtblue}{3}. \acs{CLDA} uses these projections to assess the corresponding dependencies in the \ac{LDG} \circledstep{cldaorange}{4}, which connects service requirements to monitoring witnesses, network paths, shared failure domains (e.g., routers or links whose failure could disrupt multiple witnesses), and recovery prerequisites.

Using the projected network state and \acs{LDG}, \acs{CLDA} evaluates four assurance requirements \circledstep{cldaorange}{5}. \emph{Semantic binding} verifies that each witness continues to measure the intended service property within its required network scope. \emph{Evidence independence} checks that witnesses designated as independent do not share failure domains prohibited by the assurance policy. \emph{Prefix safety} requires the applicable assurance requirements to hold after every action in a multi-step plan, rather than only in its final state. For example, moving a service before updating its latency monitor may temporarily leave the service without valid monitoring, even if the completed plan restores it. \emph{Executable recovery} checks that the resources needed for registered rollback operations remain available. Based on these checks, \acs{CLDA} accepts, repairs, or rejects the plan \circledstep{cldaorange}{6}. Plans satisfying the requirements, either as proposed or following a supported make-before-break repair, are authorized for execution through the network controller.

\textbf{Network State and Twin Projection.} The \acs{NDT} uses observed routes, routing policies, interface status, and performance measurements to represent the current network. Its what-if analysis predicts the effects of candidate changes before they are applied, including bidirectional reachability, path quality, and network-level objectives. However, an acceptable projected outcome does not necessarily mean that monitoring still represents the service or that recovery remains possible.

\textbf{Agentic Planning.} The \acs{LLM}-based planner translates an operator's intent into a sequence of proposed network actions and can revise the plan based on validation feedback. Consistent with emerging supervision architectures that place \acs{AI}-driven planning within a broader framework for observability, control, and intervention~\cite{ma2026icon}, the planner does not independently authorize or execute its proposed actions.

\textbf{Loop Dependency Assurance.} \emph{Monitoring evidence} consists of measurements, such as latency, packet loss, or reachability, used to assess service requirements. A \emph{witness} is the monitoring mechanism that produces these measurements, such as an active latency probe or a service-health monitor. Its \emph{binding} specifies the service property and network scope that the measurements are intended to represent. The \acs{LDG} records the \emph{assurance dependencies} needed for continued service verification and recovery. These include relationships between service requirements and witnesses, their associated network paths, shared failure domains that may affect multiple witnesses, and \emph{recovery prerequisites} such as paths and interfaces needed to execute a rollback.

The service-assurance graph defined in RFC~9417 supports service-health assessment and fault diagnosis~\cite{claise2023sain}. In contrast, \acs{CLDA} uses the \acs{LDG} to assess whether proposed network changes preserve the monitoring and recovery capabilities needed for subsequent decisions. Network verification checks forwarding and policy invariants, while consistent-update techniques reason about intermediate network states~\cite{khurshid2013veriflow,reitblatt2012updates}. \acs{CLDA} complements these approaches by integrating dependency checks with \acs{NDT} projections, authorization, and make-before-break repair within the management loop.

Unlike the agentic \acs{AI} planner, \acs{CLDA} uses deterministic procedures rather than an \acs{LLM} for assurance decisions and repair synthesis. These procedures evaluate the projected forwarding state against the configured assurance dependencies after each planned action. For supported temporary monitoring gaps, rule-based repair synthesis generates additional witness configurations before the affected changes. Repaired plans undergo the same validation before authorization. During execution, the network state is updated after each applied action to support continued assurance checks.

\section{Preserving Assurance Across Network Changes}
\label{sec:assurance}

\begin{figure}[!t]
	\centering
	\includegraphics[width=\columnwidth]{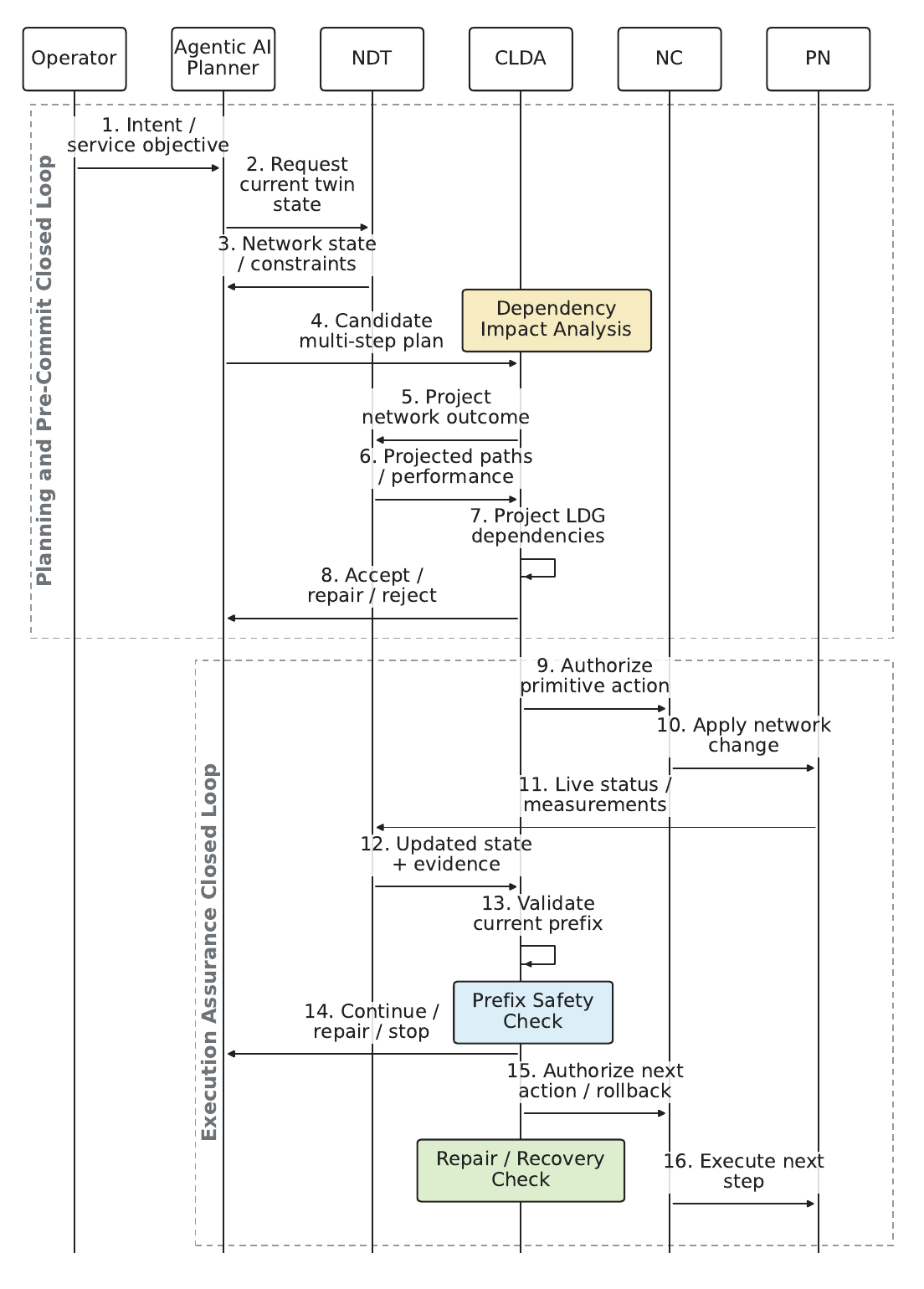}
	\caption{Planning and execution with \acs{CLDA}. Candidate actions are projected before commit; during execution, the live state is refreshed after network-changing actions so the controller can continue, repair, stop, or recover.}
	\label{fig:procedure}
\end{figure}

\acs{CLDA} targets three ways a valid-looking network change can decouple the closed loop from its managed service.

\textbf{Semantic Evidence.} A measurement may be recent and accurate without providing valid evidence for the service property it is intended to verify. A route change may move production traffic to a new branch while its round-trip latency measurements continue to measure the previous branch. Those measurements may be accurate for the old route, but no longer represent the path taken by the service. \acs{CLDA} therefore binds each service requirement to the scope of its evidence. Under an exact-path round-trip latency contract, the witness must follow the service's projected forward and return paths. A witness that remains online but no longer follows those paths is excluded from the assurance decision. This follows the operational principle that monitoring scope must track the path being assured~\cite{dalessandro2017rfc8256}.

\textbf{Independent Evidence.} Multiple witnesses provide independent confirmation only when they avoid shared failure domains. For example, rerouting can cause previously disjoint witnesses to traverse the same critical router, tunnel, or collector, weakening the assurance they provide. \acs{CLDA} hence projects the resources on which each witness depends and rejects a plan when the resulting sharing violates the declared independence policy. This complements digital-twin conflict checking, which focuses on whether network applications or policies remain mutually consistent~\cite{polverini2026conflicts}.

\textbf{Safe Execution and Recovery.} A multi-step plan can end safely even though an intermediate action removes required evidence. Likewise, a registered rollback is useful only while the paths and interfaces needed to execute it remain available. \acs{CLDA} therefore checks the projected network and dependency state after every resulting execution prefix, i.e., the intermediate result. A prefix is admissible only if it preserves the witness bindings, declared independence constraints, and rollback prerequisites. This extends established reasoning about intermediate network updates~\cite{reitblatt2012updates} to modeled monitoring and recovery dependencies. 

When a temporary monitoring gap is the only violation and a repair is available, \acs{CLDA} inserts temporary witnesses before the affected action and removes them once permanent coverage is restored. Other violations, such as semantic mismatch, independence collapse, and unusable rollbacks, are rejected rather than automatically rewritten.

Fig.~\ref{fig:procedure} illustrates the planning and execution workflow. During planning and pre-commit (Steps 1--8), the operator provides an intent and service objective to the agentic \acs{AI} planner (Step 1). The planner requests the current network state from the \acs{NDT} (Step 2), which returns the network state and constraints (Step 3). Using this information, the planner generates a candidate multi-step plan and submits it to \acs{CLDA} (Step 4). \acs{CLDA} requests an \acs{NDT} projection of the proposed changes (Step 5) and receives their expected effects on network paths and performance (Step 6). It then projects the corresponding \acs{LDG} dependencies (Step 7) and determines whether the plan can be accepted, repaired, or must be rejected (Step 8). During execution (Steps 9--16), \acs{CLDA} authorizes an individual network action (Step 9), which the network controller applies to the \ac{PN} (Step 10). 

The resulting network status and measurements are collected by the \acs{NDT} (Step 11), which provides the updated state and monitoring evidence to \acs{CLDA} (Step 12). \acs{CLDA} validates the resulting execution prefix against the assurance requirements (Step 13) and reports whether execution can continue, requires repair, or must stop (Step 14). If the next operation is permitted, \acs{CLDA} authorizes the subsequent action or an applicable rollback (Step 15), which the network controller executes (Step 16). This process repeats after each network-changing action.

\section{Case Study: Dependency-Aware Autonomous Reconfiguration}
\label{sec:case_study}

We evaluate \acs{CLDA} using controlled network changes, LLM-generated plans, and live testbed execution. The study examines whether acceptable projected network outcomes can mask violations of assurance dependencies, whether \acs{CLDA} can prevent such violations during a multi-step reconfiguration, and whether the same protection applies to plans generated by an agentic \acs{AI} planner.

\subsection{Prototype and Controlled Scenarios}
The eight-node Containerlab testbed consists of four Linux routers (r1--r4), a client, a server, and two monitoring hosts connected to r2 and r3 to provide evidence of reachability. Router r1 serves as the ingress, and router r4 as the egress, while r2 and r3 provide parallel transit paths. Initially, the service traverses r1--r2--r4 in the forward direction and r4--r2--r1 in the return direction. The client uses separate source IP addresses for service traffic and its latency witness, which provides round-trip measurements representing the service path. Linux policy-routing rules allow service traffic to be rerouted through r3, while its latency witness continues measuring the original path through r2. During supported make-before-break repairs, \acs{CLDA} provisions temporary witnesses to maintain monitoring coverage while the permanent witness is reconfigured. The observed network states enable reconstruction of the forward and return paths for service and monitoring traffic, allowing for round-trip time and packet-loss measurements. The evaluated \acs{NDT} uses this information to assess the projected effects of candidate changes on bidirectional reachability, path quality, and operational objectives.

\acs{CLDA} combines the observed network state with witness binding, failure-domain independence, and recovery prerequisites before authorizing an action. We compare \acs{CLDA} with the \acs{NDT} preflight validation described above, as well as three additional baselines defined for this study. The freshness/availability baseline checks whether required witnesses remain reachable and provide measurements, without verifying that they represent the service. The narrow static policy blocks changes to routing entries used by designated monitoring witnesses, whereas the broad static policy blocks all route and interface changes on the four routers. These illustrative baselines are not components of \acs{CLDA}, but allow us to compare dependency-aware authorization with simpler checks and fixed restrictions that may either overlook assurance violations or reject permissible changes.

To evaluate how network changes can compromise different assurance requirements, we construct four controlled scenarios, each targeting a distinct dependency violation. In semantic-binding (S1), the service forward path passes through r3 while its latency witness remains on r2. In independence-collapse (S2), two monitoring probes test network reachability through different paths. Rerouting one probe causes both to depend on a shared router, violating the independence policy and risking disruption of both measurements if that router fails. In recovery-erosion (S3), service and witness traffic migrate to r3 before the original r1--r2 ingress route is removed, making the registered rollback through r2 unusable. Finally, in temporal-prefix (S4), production traffic moves to r3 before its permanent witness is moved. Unlike S1--S3, S4 reaches a final configuration that satisfies the assurance requirements, but its intermediate execution states lack the required evidence.

S1 illustrates why fresh measurements are not necessarily valid evidence. In the pre-commit experiment, the evaluated \acs{NDT} accepts the projected reroute through r3, while \acs{CLDA} rejects it because the contracted latency witness would remain on r2. In a separate live experiment, we introduce $40\,\mathrm{ms}$ of delay on the r3 branch and reroute production traffic. The service RTT reaches $40.139\,\mathrm{ms}$, exceeding its $25\,\mathrm{ms}$ objective, while the old witness remains fresh and reports only $0.086\,\mathrm{ms}$. The r3 ground-truth probe records $40.147\,\mathrm{ms}$. Using routing state collected from the network, \acs{CLDA} detects mismatches between the service and witness paths in both directions, leaving no semantically valid witness for the latency requirement.  S2 is rejected when the witnesses lose their required independence, while S3 is rejected before executable recovery is lost. In S4, \acs{CLDA} provisions temporary witnesses before migration, preserving assurance throughout the reconfiguration.

To assess whether the evaluated approaches distinguish assurance violations from permissible network changes, we construct a targeted test set of 15 network-change plans. Five plans deliberately violate the declared assurance requirements, specifically, S1--S4, and an additional return-path binding violation in which the service's return route changes while its latency witness retains the previous return path. The ten benign plans cover service and witness refreshes, changes outside the active assurance dependencies, idempotent interface updates, a no-op, and a recovery-preserving refresh. The comparison evaluates them before any repair is applied, with each plan classified according to the expected outcome based on the assurance requirements, independently of its evaluated decisions.

\begin{figure}[!t]
\centering
\includegraphics[width=\columnwidth]{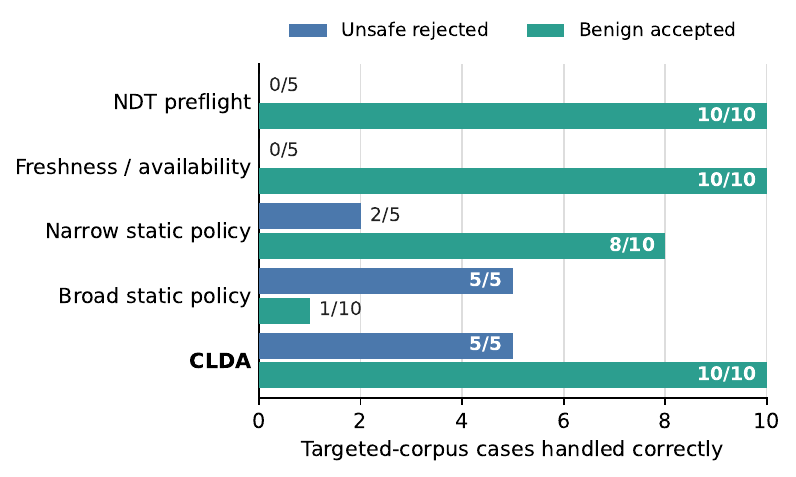}
\caption{Outcomes of the evaluated assurance approaches on $15$ targeted cases, comprising five unsafe and ten benign changes. Bars indicate the number of unsafe changes rejected and benign changes accepted, with counts relative to each category's total.}
\label{fig:tradeoff}
\end{figure}

The evaluated \acs{NDT} preflight validation and freshness/availability accept all 15 plans, including the five unsafe ones. The narrow static policy rejects two unsafe plans but also blocks two benign plans. The broad static policy rejects all five unsafe plans but blocks nine benign plans. In comparison, \acs{CLDA} rejects all five unsafe plans and accepts all ten benign plans. Fig.~\ref{fig:tradeoff} summarizes these outcomes. Because the test set was deliberately constructed to exercise the modeled assurance requirements, the results demonstrate the intended behavior under controlled conditions rather than general detection performance across operational networks.

\subsection{Agentic Planning with Qwen3.8-27B}
We next integrate a locally deployed \acs{LLM} planner that translates natural-language intents (prompts) into candidate network plans. Only plans that satisfy both \acs{NDT} and \acs{CLDA} requirements, either as proposed or following deterministic repair, are authorized for execution.

We use the 27-billion-parameter Qwen3.8-27B model~\cite{qwen2026model}, deployed locally with vLLM on an NVIDIA GH200 accelerator. The planner receives the operator intent, the current network state provided by the \acs{NDT}, and the set of permitted network actions. Each proposed plan contains at most eight steps, with one network-changing action per step. An independent goal check determines whether the plan satisfies the requested operational objective~\cite{cui2026netconfbench}. We call such plans \emph{goal-matched}. The \acs{NDT} then evaluates the projected network outcomes, while \acs{CLDA} checks the associated assurance dependencies. For supported temporary monitoring gaps, \acs{CLDA} can perform a \emph{deterministic repair} by inserting temporary witnesses before the affected change. If a goal-matched plan fails validation and cannot be authorized after a repair, the planner receives feedback and may propose up to two revisions. Each revision then undergoes the same procedure.

To assess whether \acs{CLDA} identifies assurance violations in \acs{LLM}-generated plans while permitting benign changes and supported repairs, we evaluate plans generated from varied operator intents and independently seeded runs. For each trial, the same Qwen-generated plan is evaluated against the \acs{NDT}-only baseline and the \acs{CLDA}-guarded workflow. The campaign covers eight intent families: two benign families, S1--S4, return-path binding, and contradictory objectives. Each family has three paraphrases and ten seeded runs per paraphrase, for a total of 240 trials. These trials evaluate variation across intents rather than distinct network topologies. Among the 235 schema-valid initial candidates, we identify 22 distinct combinations of network-action sequences and declared objectives after excluding differences in textual naming.

\begin{figure}[!t]
\centering
\includegraphics[width=\columnwidth]{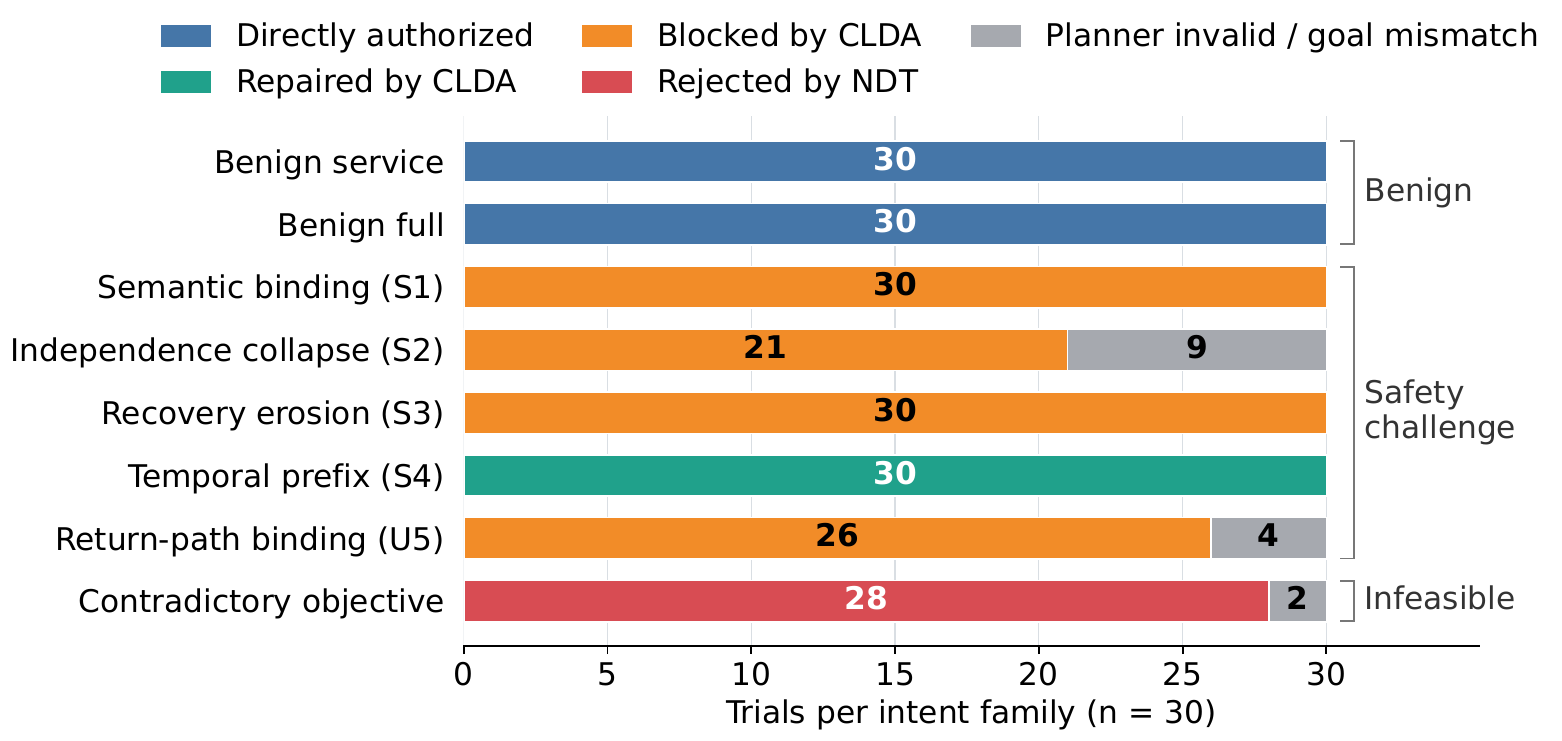}
\caption{Outcome distribution across the $240$-trial agentic planning campaign. Each bar contains $30$ trials from one intent family; planner-invalid and goal-mismatched outputs are separated from \acs{NDT} and \acs{CLDA} decisions.}
\label{fig:agentic_campaign}
\end{figure}

\begin{figure*}[!t]
    \centering
    \subfloat[\label{fig:latency_hierarchy}]{
        \includegraphics[width=0.48\textwidth]{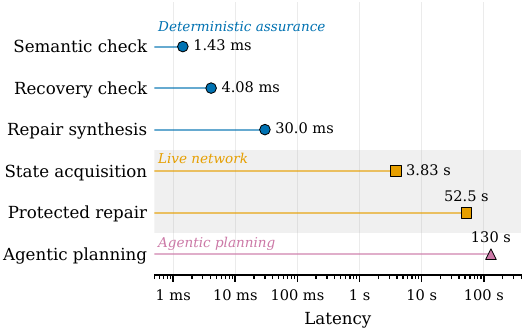}
    }
    \hfill
    \subfloat[\label{fig:scalability}]{
        \includegraphics[width=0.48\textwidth]{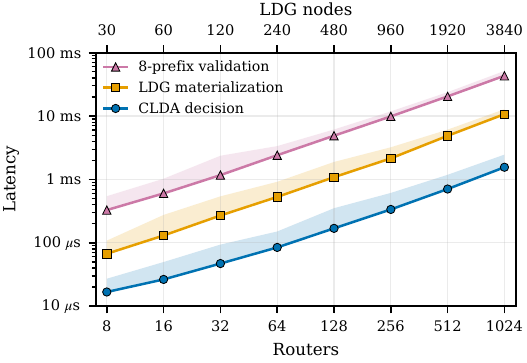}
    }
    \caption{Operational cost and scalability of \acs{CLDA}. (a) Representative prototype time scales, separating deterministic assurance from live network interaction and agentic planning. (b) Synthetic scaling from $8$ to $1{,}024$ routers ($30$ to $3{,}840$ \acs{LDG} nodes). Lines connect medians over $200$ measured repetitions after $40$ warm-ups; shaded regions extend to p95.}
    \label{fig:cost_scalability}
\end{figure*}

Return-path binding (U5) tests the complementary S1 case where the service return route changes while latency-measurement traffic still follows the original return route. Of the 240 initial plans, 235 conform to the action schema, which specifies the permitted network operations and their parameters, and 225 satisfy the independent goal check. The \acs{NDT} preflight validation rejects 28 goal-matched plans because they fail its objective-consistency check. The remaining 197 plans comprise 60 benign and 137 safety-challenge plans. All 60 benign plans are authorized by \acs{CLDA}. Among the 137 safety-challenge plans, \acs{CLDA} identifies the modeled assurance violations, repairs the 30 S4 plans by synthesizing temporary witnesses, and blocks the remaining 107. Fig.~\ref{fig:agentic_campaign} summarizes these initial-plan decisions and deterministic S4 repairs. Planner-invalid and goal-mismatched outputs are reported separately from assurance decisions.

Feedback-guided refinement was exercised in 135 initially rejected, goal-matched trials, comprising 107 blocked by \acs{CLDA} and 28 rejected by the evaluated \acs{NDT}. Across 270 revision attempts, no additional plan passed all required authorization checks. These results are specific to the evaluated intents, \acs{LLM} configuration, and network testbed. They do not indicate how frequently unsafe plans would arise in operational networks. We also execute one representative Qwen-generated S4 plan on the live testbed. Before actuation, the restricted executor revalidates the candidate and locally synthesizes the deterministic \acs{CLDA} repair. All eight executed prefixes remain \acs{CLDA}-safe, all nine recovery checks (one before execution and eight after the executed prefixes) succeed, the intended r3 service and witness paths are reached, and cleanup restores the r2 baseline. This run demonstrates the integration of LLM planning, deterministic assurance, repair, and live execution for a representative generated plan.

\subsection{Operational Cost and Live Execution}
We evaluate the computational cost of deterministic assurance separately from live network operations and \acs{LLM} planning. Core decision checks measure the time required to evaluate candidate changes against the available network state and assurance dependencies, excluding network measurements and \acs{LLM} inference. Repair synthesis measures the computational cost of generating temporary witness actions without executing them. Live-state acquisition measures the time required to retrieve routing information and probe measurements from the running network. Finally, protected execution measures the end-to-end time to apply a multi-step plan and validate the network state after each action. We perform 200 repetitions per case for core decision checks, 100 for repair synthesis, and 30 trials each for live-state acquisition and protected execution.

Using the same previously collected network state, we measure the additional decision time introduced by \acs{CLDA} relative to the freshness/availability baseline. Across 13 cases not involving recovery validation, the mean overhead is $12.21\,\mu\mathrm{s}$, with a median of $11.65\,\mu\mathrm{s}$ and a 95th percentile of $27.43\,\mu\mathrm{s}$, calculated across the per-case means. For the two cases requiring recovery validation, the mean additional overhead increases to $1.08\,\mathrm{ms}$, ranging from $0.885$ to $1.277\,\mathrm{ms}$, as \acs{CLDA} checks whether the registered rollback remains executable.

Fig.~\ref{fig:cost_scalability}(a) compares the execution times of deterministic assurance checks, repair synthesis, live network-state acquisition, and \acs{LLM}-based planning. Representative semantic-binding (S1) and recovery (S3) checks take $1.43\,\mathrm{ms}$ and $4.08\,\mathrm{ms}$, respectively. These are absolute decision times, rather than the additional \acs{CLDA} overhead relative to the freshness/availability baseline reported above. Synthesizing the S4 repair takes approximately $30\,\mathrm{ms}$, while acquiring live routing and measurement state takes $3.83\,\mathrm{s}$. Unlike the agentic planner, \acs{CLDA} uses deterministic, rule-based checks over the projected network state and \acs{LDG}, without invoking an \acs{LLM} for assurance decisions or repair synthesis.

Across all 240 planning trials, the mean total \acs{LLM} planning time is $130.47\,\mathrm{s}$ per trial, including feedback-guided revisions when required. The 105 trials without feedback-guided replanning average $36.37\,\mathrm{s}$, whereas the 135 trials involving two revisions average $203.66\,\mathrm{s}$. The average total number of input tokens sent to the model per trial also increases from approximately $2{,}992$ to $11{,}014$. The longer planning times, therefore, reflect additional \acs{LLM} inference during replanning, rather than the computational cost of \acs{CLDA} assurance checks.

Finally, we execute the deterministically repaired S4 sequence 30 times, separately from the Qwen-generated planning demonstration. All 30 executions succeed, with all 8, per-run, execution prefixes satisfying the modeled \acs{CLDA} requirements (240 prefixes in total). The complete protected S4 execution averages $52.46\,\mathrm{s}$ across eight execution prefixes, including network changes, live-state acquisition, and assurance validation after each action.

\textbf{Scalability.} Fig.~\ref{fig:cost_scalability}(b) reports how deterministic assurance time changes as the modeled network grows. Because the live testbed is intentionally compact to expose the assurance failures under controlled execution, we also evaluate the deterministic assurance core separately on a synthetic network ranging from 8 to 1,024 routers. The number of services, witnesses, postconditions, recovery contracts, and resources grows proportionally, expanding the \acs{LDG} from 30 nodes and 34 edges to 3,840 nodes and 4,352 edges. Each configuration uses 40 warm-up runs and 200 measured repetitions pinned to one logical core of an Intel Core i3-1115G4 at 3.0\,GHz.

With dependency state already instantiated, median \acs{CLDA} decision latency increases from $16.74\,\mu\mathrm{s}$ at 8 routers to $1.56\,\mathrm{ms}$ at 1,024 routers, with a $2.46\,\mathrm{ms}$ p95 at the largest configuration. Materializing the corresponding \acs{LDG} has median latencies of $67.13\,\mu\mathrm{s}$ and $10.72\,\mathrm{ms}$, respectively. A log-log fit of \acs{CLDA} decision latency gives an exponent of $0.94$ ($R^2=0.994$), consistent with near-linear scaling over the evaluated range. The median time for complete validation of an eight-prefix plan increases from $0.33\,\mathrm{ms}$ to $43.24\,\mathrm{ms}$, with a fitted exponent of $1.01$ ($R^2=0.999$).

We separately fix the dependency context at 256 routers and 960 \acs{LDG} nodes and increase plan depth from 1 to 32 execution prefixes. Median validation time increases from $7.53\,\mathrm{ms}$ to $19.59\,\mathrm{ms}$; a linear fit yields $R^2=0.996$. These results indicate that, for the dependency structures evaluated here, deterministic assurance remains lightweight as both dependency context and plan depth increase.
\section{Open Research Issues}
\label{sec:open_issues}

\textbf{Learning and Maintaining Dependency Context.} The \acs{LDG} used in the eight-node evaluation is small and route-centric. Larger deployments may also need to represent overlays, tunnels, telemetry pipelines, service functions, controllers, credentials, and shared infrastructure. Future systems will need incremental graph maintenance and dependency discovery from topology, configuration, telemetry, and controller metadata. The current prototype combines live forwarding and measurement state with declared assurance requirements. At a larger scale, the challenge is therefore not only to check the resulting \acs{LDG}, but also to establish the provenance, freshness, and confidence of its relations as services, telemetry pipelines, and recovery mechanisms evolve.

\textbf{Continuous State Synchronization.} Obtaining the current routing state and active measurements takes much longer than the deterministic assurance checks in the evaluated testbed. Sub-second loops will require event-driven telemetry, streaming topology updates, and selective evidence acquisition so assurance-relevant state is maintained continuously rather than rediscovered after each action. The scalability experiment isolates deterministic assurance and does not establish how generative planning scales with network size. Larger deployments may increase both the context presented to the planner and the candidate action space, motivating context selection or retrieval and bounded replanning while deterministic assurance remains on the authorization path. A key systems question is how much dependency context can be cached without making authorization stale.

\textbf{Cross-Domain Security and Privacy.} The \acs{LDG} can reveal sensitive information such as monitor placement, recovery paths, and shared critical dependencies. Production \acp{CDT} therefore require access control, integrity protection, and selective disclosure for assurance metadata that crosses organizational boundaries. Policies must also express acceptable failure-domain sharing and recovery guarantees for services with different risk levels.

\textbf{Uncertainty and Multi-Agent Coordination.} Real deployments will face incomplete failure-domain information, uncertain telemetry, and multiple agents modifying related resources. Confidence and provenance on \acs{LDG} relations could identify actions that need additional evidence or human escalation. Multi-agent systems also need coordination so that one plan cannot invalidate another agent's assurance context between projection and commit.

\section{Conclusion}
\label{sec:conclusion}
Network changes proposed by an agentic planner can satisfy an immediate objective while weakening the evidence or recovery mechanisms needed for subsequent closed-loop decisions. \acs{CLDA} addresses this self-decoupling risk by extending \acs{NDT} what-if analysis to the dependencies that sustain assurance. The \acs{LDG} enables semantic binding, evidence independence, prefix safety, executable recovery, and make-before-break repair before a plan is committed. Our case study shows that these dependency failures can escape outcome-oriented validation while deterministic assurance can expose or repair them with modest decision overhead. Dependency-aware assurance extends closed-loop validation beyond the final network state to the actions used to reach it.

\balance
\bibliographystyle{IEEEtran}
\bibliography{inputs/references}

\vspace{-33pt}
\begin{IEEEbiographynophoto}{Ahmed~Hussain}
is a Doctoral Researcher at KTH Royal Institute of Technology, Stockholm, Sweden. His research interests include IoT security and privacy, wireless security, Edge AI, Agentic AI, and AI-enabled cybersecurity. Website: \url{https://ahmed-hussain.net}.
\end{IEEEbiographynophoto}
\vspace{-33pt}
\begin{IEEEbiographynophoto}{Konstantinos~Kalogiannis} is a PhD Candidate at KTH Royal Institute of Technology, Stockholm, Sweden. His research interests include security and privacy in cyber-physical systems, with a focus on vehicular ad hoc networks (VANETs) and platooning, and the security analysis of large-scale software systems.
\end{IEEEbiographynophoto}
\vspace{-33pt}
\begin{IEEEbiographynophoto}{Salahuddin~Salahuddin}
is currently an AI Researcher at SSH Communications Security. He received his M.Sc. through a double-degree program from Aalto University, Finland, and KTH Royal Institute of Technology, Sweden. His research interests include artificial intelligence, cybersecurity, machine learning, and the application of AI techniques to security-related challenges.
\end{IEEEbiographynophoto}
\vspace{-33pt}
\begin{IEEEbiographynophoto}{Hexu~Li} is currently a Researcher working with the Tech Validation – AI department at the Chief Technology Officer (CTO) Organization, Lenovo, Beijing, China. He received his Master's degree in Communication Systems from KTH Royal Institute of Technology, Stockholm, Sweden. 
\end{IEEEbiographynophoto}
\vspace{-33pt}
\begin{IEEEbiographynophoto}{Thiemo~Voigt} received the Ph.D. degree from Uppsala University, Sweden, in 2002. He is currently a Full Professor at the Department of Electrical Engineering, Uppsala University, and a Senior Researcher at RISE Computer Science. His current research interests include low-power networking, system software, and machine learning for embedded networked devices and the Internet of Things, as well as in-body communication and 6G Ambient IoT.
\end{IEEEbiographynophoto}

\end{document}